\documentclass[letterpaper, 10 pt, conference]{IEEEtran}

\IEEEoverridecommandlockouts

\usepackage{makecell}
\usepackage{subcaption}
\usepackage{pgf}
\usepackage{url}
\usepackage{hyperref}

\usepackage{comment}
\usepackage{cite}
\usepackage{amsmath,amssymb,amsfonts}
\usepackage{algorithmic}
\usepackage{graphicx}
\usepackage{textcomp}
\usepackage{xcolor}
\usepackage[font={footnotesize}]{caption}
\usepackage{colortbl,xcolor} 

\def\BibTeX{{\rm B\kern-.05em{\sc i\kern-.025em b}\kern-.08em
    T\kern-.1667em\lower.7ex\hbox{E}\kern-.125emX}}
\begin{document}

\title{\LARGE \bf Towards Versatile Opti-Acoustic Sensor Fusion and 
Volumetric Mapping
for Safe Underwater Navigation

\thanks{{\footnotesize{$^{1}$I. Collado-Gonzalez, and B. Englot are with Stevens Inst. of Technology, Hoboken, NJ, USA, \{\texttt{icollado}, \texttt{benglot}\}\texttt{@stevens.edu}. $^{2}$J. McConnell is with the U.S. Naval Academy, Annapolis, MD, USA, \texttt{jmcconne@usna.edu}. \textbf{This research was supported by grants NSF IIS-1652064, USDA-NIFA 2021-67022-35977, and ONR N00014-24-1-2522.} The views expressed in this paper are those of the author(s) and do not reflect the official policy or position of the U.S. Naval Academy, Department of the Navy, the Department of War, or the U.S. Government.}}}
}

\author{\IEEEauthorblockN{Ivana Collado-Gonzalez$^{1}$, John McConnell$^{2}$, and Brendan Englot$^{1}$}}

\maketitle

\begin{abstract}
Accurate 3D volumetric mapping is critical for autonomous underwater vehicles operating in obstacle-rich environments. Vision-based perception provides high-resolution data but fails in turbid conditions, while sonar is robust to lighting and turbidity but suffers from low resolution and elevation ambiguity. This paper presents a volumetric mapping framework that fuses a stereo sonar pair with a monocular camera to enable safe navigation under varying visibility conditions. Overlapping sonar fields of view resolve elevation ambiguity, producing fully defined 3D point clouds at each time step. The framework identifies regions of interest in camera images, associates them with corresponding sonar returns, and combines sonar range with camera-derived elevation cues to generate additional 3D points. Each 3D point is assigned a confidence value reflecting its reliability. These confidence-weighted points are fused using a Gaussian Process Volumetric Mapping framework that prioritizes the most reliable measurements. Experimental comparisons with other opti-acoustic and sonar-based approaches, along with field tests in a marina environment, demonstrate the method’s effectiveness in capturing complex geometries and preserving critical information for robot navigation in both clear and turbid conditions. Our code is open-source to support community adoption.
\end{abstract}

\vspace{-1mm}
\section{Introduction}
\vspace{-1mm}
Autonomous Underwater Vehicles (AUVs) are essential for tasks such as maintenance, mapping, and exploration, often in turbid waters. Information-gathering tasks often require large-scale environment mapping, while maintenance tasks demand accurate close-range reconstruction to guide manipulators and avoid unmodeled obstacles. Reliable sensing and mapping across diverse conditions and scales is therefore critical for AUV autonomy and safety.

Vision-based perception has shown partial success underwater \cite{Yang2021}
, but its reliance on salient features makes it sensitive to illumination changes, blurring, and halo effects, resulting in unreliable performance in scattering media. Sonar sensors provide robustness in turbid water and under variable lighting, yet they have inherent limitations: profiling sonars capture narrow slices, 3D sonars are emerging \cite{Sonar3D} but remain range-limited, and imaging sonars produce 2D projections of 3D scenes, restricting vertical Field Of View (FOV) and introducing elevation ambiguity. Despite advances, high-resolution 3D sonar reconstruction remains challenging. Sonar’s low resolution favors large-scale mapping but limits detection of fine features and close-range objects critical for collision avoidance and manipulation.
\begin{figure}[t]
    \centering
    \begin{subfigure}{0.5\columnwidth}
        \centering
        \includegraphics[width=\linewidth]{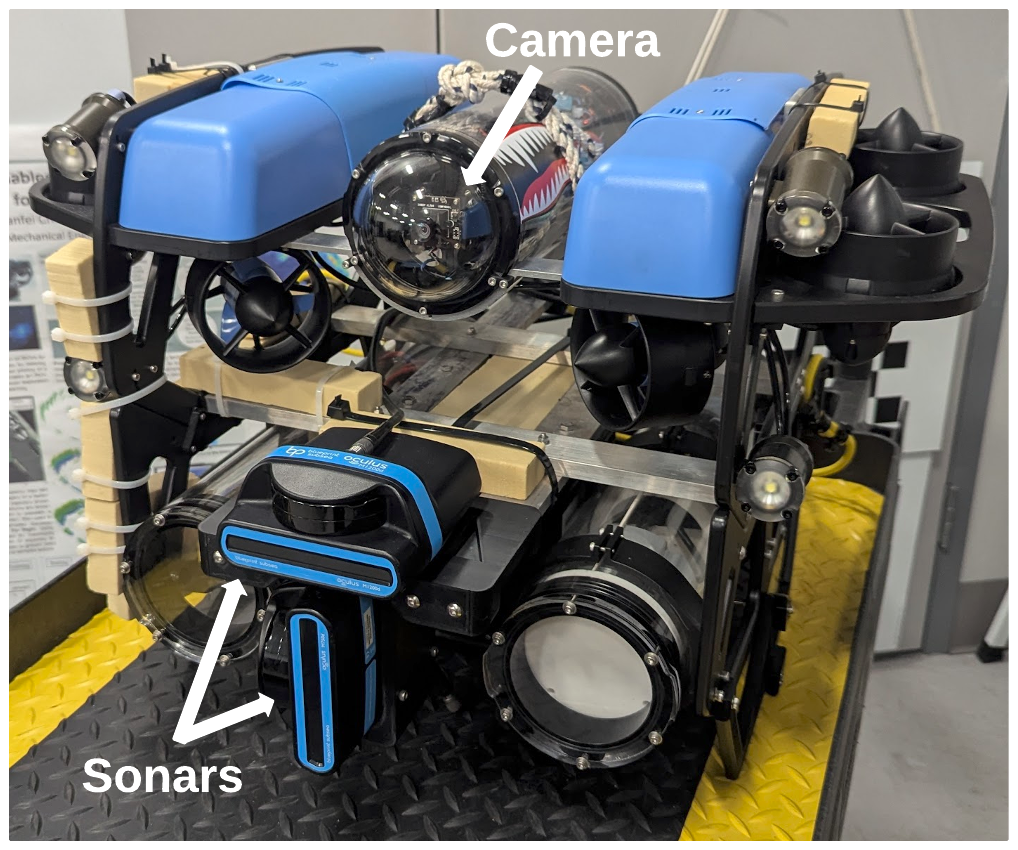} 
        \caption{BlueROV2 platform.}
        \label{fig:ROV}
    \end{subfigure}
    \hfill
    \begin{subfigure}{0.48\columnwidth}
        \centering
        \includegraphics[width=\linewidth]{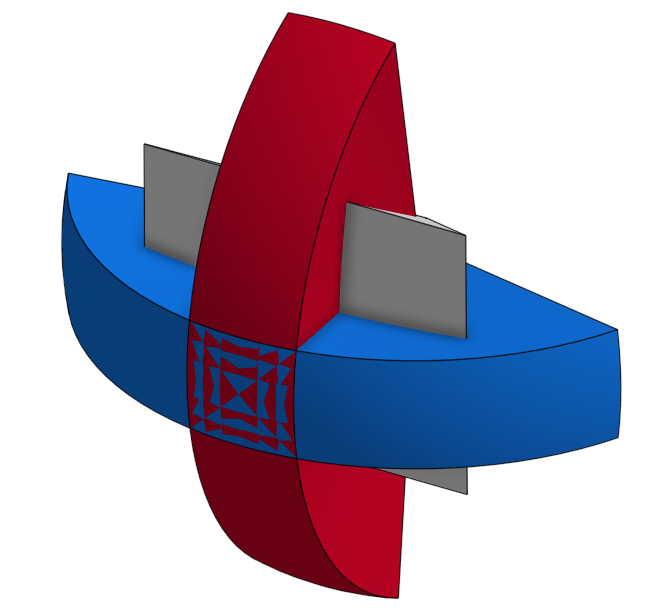} 
        \caption{Fields of view of sensors.}
        \label{fig:FOV}
    \end{subfigure}
    \begin{subfigure}{\columnwidth}
        \centering
        \includegraphics[width=\columnwidth]{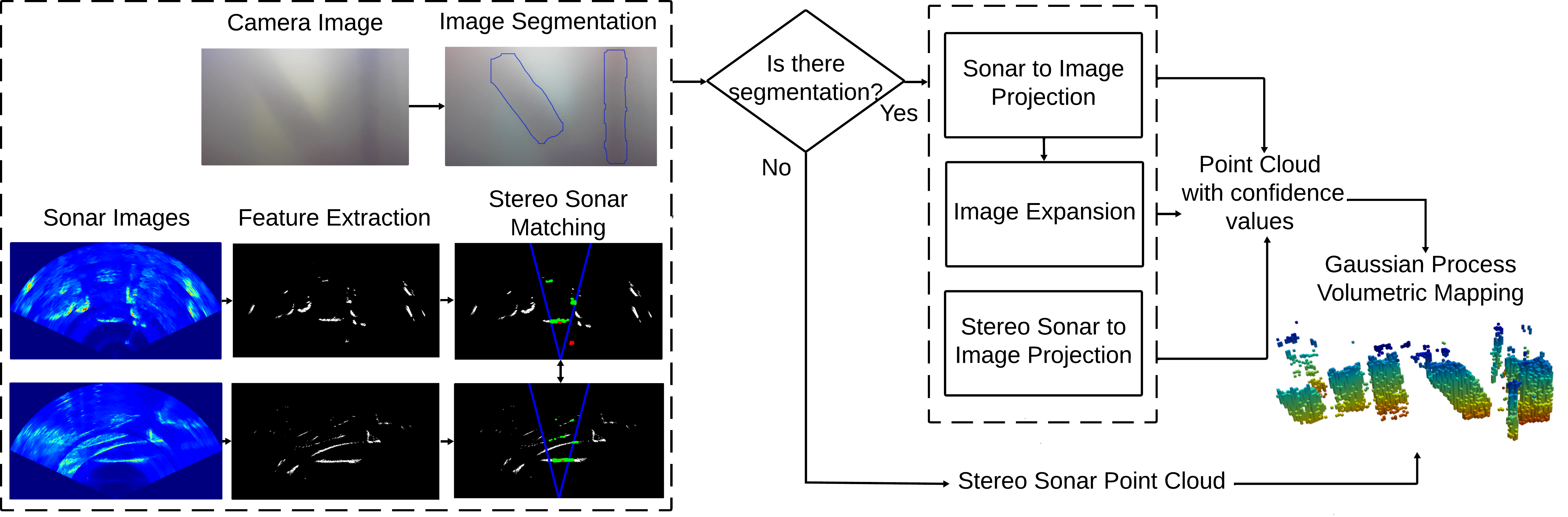} 
        \caption{Proposed sensor fusion and mapping pipeline.}
        \label{fig:pipeline}
    \end{subfigure}
    \caption{{\textbf{Overview.} (a) Our BlueROV2 platform; (b) an illustration of the fields of view of its camera and sonars, with the gray volume representing the camera and the blue/red volumes indicating the horizontal/vertical sonars respectively; (c) an illustration of our proposed volumetric mapping pipeline.}}
    \label{fig:FOVandROV}
    \vspace{-4mm}
\end{figure}

These limitations motivate the fusion of complementary sensing modalities. Optical imaging provides high-resolution detail, while acoustic imaging ensures robustness in turbid waters and extended range. Fusion can enhance 3D reconstruction, but differing sensor image-formation mechanisms across sensors complicate direct feature matching. Most approaches treat visual cues as primary input and use sonar for refinement \cite{Hu2023}, which fails in turbid environments. A recent opti-acoustic method \cite{Collado2025} builds on sonar features, enhancing reconstructions with visual data to expand vertical FOV. Although it removes the dependency on reliable visual features, its application is limited to simple object geometries. These challenges highlight the need for a fusion strategy that prioritizes sonar while flexibly incorporating visual data for reliable 3D mapping in complex underwater scenes.

This work addresses underwater sensing and mapping under variable visibility conditions. This method leverages overlapping sensor FOVs to compensate for missing information in both camera and sonar, providing complementary distance and elevation cues. Confidence values derived from the fused 3D solution guide the mapping process by weighting measurements according to reliability. The framework relies primarily on sonar while incorporating camera data only when available. A CNN-based segmentation module is trained efficiently using transfer learning and data augmentation on visual data, avoiding the need for large sonar datasets. This approach generates scene maps from a single pose, eliminating the need for specific motion patterns, and focuses on extracting information relevant for safe navigation. By addressing the limitations of existing approaches, this work improves AUV performance in complex underwater environments.

The key contributions of this paper are:
\begin{itemize}
\item To the best of our knowledge, this is the first underwater sensing strategy to fuse information from a stereo sonar pair and a monocular camera.
\item A confidence-based volumetric mapping framework that prioritizes reliable measurements to preserve relevant information for safe robot navigation.
\item Experimental comparisons in a tank environment demonstrate the accuracy of the proposed framework at mapping complex geometries.
\item A real-world experiment showing  our framework's adaptability to low-visibility conditions.
\item Public release of our code and data to facilitate reproducibility:\\ 
\footnotesize\textbf{\href{https://github.com/ivanacollg/stereosonar_camera_mapping}{https://github.com/ivanacollg/stereosonar\_camera\_mapping}}

\end{itemize}

The remainder of the paper is structured as follows. Sec. \ref{sec:related_works} reviews related work, Sec. \ref{sec:problem_formulation} presents the problem formulation, Sec. \ref{sec:proposed_approach} details the proposed method, Sec. \ref{sec:results} presents experiments and results, and Sec. \ref{sec:conclusion} concludes the paper.

\vspace{-1mm}
\section{Related Works}
\label{sec:related_works}
\textbf{Volumetric Mapping:}
Volumetric occupancy mapping is an active research area \cite{Gomes2023, Grimaldi2024FRAGG, Jung2025, Cai2024}, with voxelized maps widely used for real-time 3D planning \cite{Edabi2024Darpa, Lindqvist2024}. Above-water range sensors, such as LiDAR and RGB-D cameras, often provide dense, accurate, and evenly distributed measurements, making them well suited for signed distance surface reconstructions, such as Truncated Signed Distance Field (TSDF)-based mapping, which exploit dense data to produce continuous surfaces. In contrast, underwater sensing presents additional challenges: sonar returns are typically sparse and noisy, making probabilistic mapping approaches especially valuable. Consequently, occupancy-based methods, which explicitly model uncertainty, remain highly relevant for underwater applications.

Occupancy grid mapping represents the environment by maintaining an independent probability of occupancy for each cell, updated incrementally using Bayesian filtering. OctoMaps \cite{Octomap2013} improve scalability through octree structures, but they assume statistical independence between cells and update only those intersected by beams. As a result, sparse measurements can produce discontinuous maps, which may lead to unsafe path planning if gaps are misclassified as free space.

To address these limitations, implicit mapping approaches represent occupancy as a continuous distribution. Gaussian Process (GP) occupancy mapping models spatial correlations between cells, enabling probabilistic inference in unobserved regions \cite{OCallaghan2012}. Unlike grid-based methods, GPs can extrapolate across sparse data, filling sensor gaps with correlated occupancy estimates and improving classification accuracy, while GP kernel design allows balancing conservative versus aggressive predictions. 
\color{black}The major drawback of GP regression is the high computational complexity. Wang presented an improved formulation of GP occupancy mapping using nested Bayesian
Committee Machines (BCMs) and test data, which reduces computational complexity and enables real-time 3D mapping \cite{WangJinkun2016}.
\color{black}
More generally, Gaussian Processes have also been applied to signed distance fields \cite{Wu2025}, demonstrating their suitability for volumetric mapping tasks.

 \textbf{Sonar-based methods:} 
Sonars are widely used underwater because they are unaffected by lighting conditions and turbidity. Forward-looking multibeam imaging sonars provide wide Fields Of View (FOV) for navigation and collision avoidance, but they produce 2D projections of 3D scenes, without elevation information reconstruction is difficult. Several methods estimate missing elevation: Aykin \cite{Aykin2013} and Westman \cite{Westman2019} use shadow edges but assume smooth, monotonic surfaces; space-carving approaches \cite{Aykin2017, Guerneve2018} and multi-view reconstructions \cite{Park2023, WangY2022} rely on observations from multiple angles, which are often infeasible in constrained or close-range scenarios.

Learning-based methods attempt to overcome these limits. Wang \cite{WangY2021} and DeBortoli \cite{DeBortoli2019Elevatenet} infer elevation from sonar images, while Qadri \cite{Qadri2023Neusis} and Sethuraman \cite{Sethuraman2025} propose neural implicit and Gaussian splatting approaches. However, all require sonar datasets, which remain scarce. To reduce reliance on data or multiple images, McConnell \cite{McConnell2020StereoSonar} introduced stereo sonar, using orthogonally mounted sonars to jointly observe scenes without geometric assumptions or prior training, although limited by a small overlapping FOV. Later extensions \cite{McConnell2021, McConnell2025} expanded the FOV via Bayesian prediction but again required sonar training data.

\textbf{Opti-Acoustic approaches:}
Combining sonar and camera data offers a way to resolve the depth and elevation ambiguities of each modality, but fusion is challenging due to their fundamentally different sensing mechanisms. Kim \cite{Kim2019} proposed building separate optical and acoustic volumetric models that are iteratively merged, but the method is computationally expensive and not real-time. Roznere \cite{Roznere2020} and Cardaillac \cite{Cardaillac2023} matched sonar points to optical images to resolve scale ambiguity in visual Simultaneous Localization And Mapping (SLAM), but both approaches still depend on visual feature tracking, which is unreliable in turbid conditions.

Learning-based methods have also been explored: Qu \cite{Qu2024} applied Gaussian splatting to acoustic–optical reconstruction, while Qadri \cite{Qadri2024} introduced AONeuS, a neural rendering framework for sonar–camera fusion. These \color{black}techniques \color{black}achieve promising results but require sonar datasets, which are scarce.

Other approaches focus on improving feature matching robustness in poor visibility. Babaee \cite{Babaee2013} used contours, and Spears \cite{Spears2014} restricted the feature search space to bounded optical regions, though both still rely on point features degraded by turbidity. Gutnik \cite{Gutnik2024} avoided this limitation by detecting areas of interest in optical images and matching them directly to sonar depth values. Collado-Gonzalez \cite{Collado2025} proposed an optical area-of-interest to sonar matching algorithm capable of reconstructing simple geometries in both clear and turbid conditions. However, this method assumes constant object depth along the vertical axis, and its segmentation process requires parameter tuning, limiting its generality.

\vspace{-1mm}
\section{Problem formulation}
\label{sec:problem_formulation}
This work addresses the problem of \textbf{sensing and mapping} an underwater environment to support safe planning and decision-making in obstacle-rich conditions under varying visibility conditions. 
\subsection{Gaussian Process Occupancy Mapping}
We assume robot poses are known, so the occupancy probability of a map cell $m_j$ is
\begin{equation}
p(m_{\color{black}j\color{black}} \mid z_{1:t}, x_{1:t}),
\end{equation}
where $z_{1:t}$ is the set of sensor observations and $x_{1:t}$ the set of robot poses. Gaussian Process (GP) occupancy mapping formulates mapping as a continuous probabilistic regression problem. The observation model is defined as
\begin{equation}
y = f(\mathbf{x}) + \epsilon, \qquad \epsilon \sim \mathcal{N}(0, \sigma_n^2),
\label{eq:observation_model}
\end{equation}
where $\mathbf{x} \in \mathbb{R}^3$ denotes an input location, $y \color{black}\in \{0,1\}$ \color{black}is the observed latent occupancy, and $\epsilon$ is zero-mean Gaussian noise. 
A GP prior is placed over $f$:
\begin{equation}
f(\mathbf{x}) \sim \mathcal{GP}(0, k(\mathbf{x}, \mathbf{x}')),
\end{equation}
with covariance function $k(\mathbf{x},\mathbf{x}')$. \color{black}Sensor observations are used as training data \color{black}
$\{\mathbf{X}, \mathbf{y}\} = \{(\mathbf{x}_i, y_i)\}_{i=1}^N$, 
\color{black}while map cells correspond to query (test) locations\color{black}. The predictive distribution for a test location $\mathbf{x}_*$ is Gaussian
\begin{equation}
\begin{aligned}
f(\mathbf{x}_*) \mid &\mathbf{X}, \mathbf{y} \sim \mathcal{N}(\mu_*, \sigma_*^2), \\
\mu_* &= k_*^\top (K + \sigma_n^2 I)^{-1} \mathbf{y}, \\
\sigma_*^2 &= k(\mathbf{x}_*, \mathbf{x}_*) - k_*^\top (K + \sigma_n^2 I)^{-1} k_*,
\end{aligned}
\end{equation}
where $K_{i\color{black}i'\color{black}} = k(\mathbf{x}_i, \mathbf{x}_{\color{black}i'\color{black}})$, $k_* = \big[ k(\mathbf{x}_1, \mathbf{x}_*), \dots, k(\mathbf{x}_{\color{black}N\color{black}}, \mathbf{x}_*) \big]^\top$. This formulation enables smooth estimation of occupancy probabilities.

To capture real-world sharp variations, we adopt the Matérn covariance function with smoothness parameter $\nu = 3/2$:
\begin{equation}
k(d) = \sigma_f^2 \left(1 + \frac{\sqrt{3}d}{l}\right)\exp\left(-\frac{\sqrt{3}d}{l}\right),
\label{eq:matern}
\end{equation}
where $d = \|\mathbf{x} - \mathbf{x}'\|$, $\sigma_f^2$ the prior signal variance, and $l$ the length-scale. 

Regression outputs are mapped to occupancy probabilities using the logistic function:
\begin{equation}
p(y=1|\mathbf{x}_{\color{black}j\color{black}}) = \frac{1}{1 + \exp(-\gamma \omega_{j})}, 
\quad \omega_{\color{black}j\color{black}} = \frac{\sigma^2_{\min}\mu_j}{\sigma_{j}^2},
\label{eq:logistic}
\end{equation}
where $\sigma^2_{\min}$ is the minimum variance and $\gamma > 0$ is a scaling constant. Cells are classified as free, occupied, or unknown:
\begin{equation}
\text{state} =
\begin{cases}
\text{free}, & p < p_\text{free}, \; \sigma_{\color{black}j\color{black}}^2 < \sigma_t^2 \\
\text{occupied}, & p > p_\text{occupied}, \; \sigma_{\color{black}j\color{black}}^2 < \sigma_t^2 \\
\text{unknown}, & \text{otherwise}
\end{cases}
\label{eq:classification}
\end{equation}
with $\sigma_t^2$ serving as a variance threshold that expresses prediction confidence and balances the trade-off between predictive richness and classification accuracy.

\subsection{Sonar Model}
\begin{figure}
    \centering
    \includegraphics[width=0.9\linewidth]{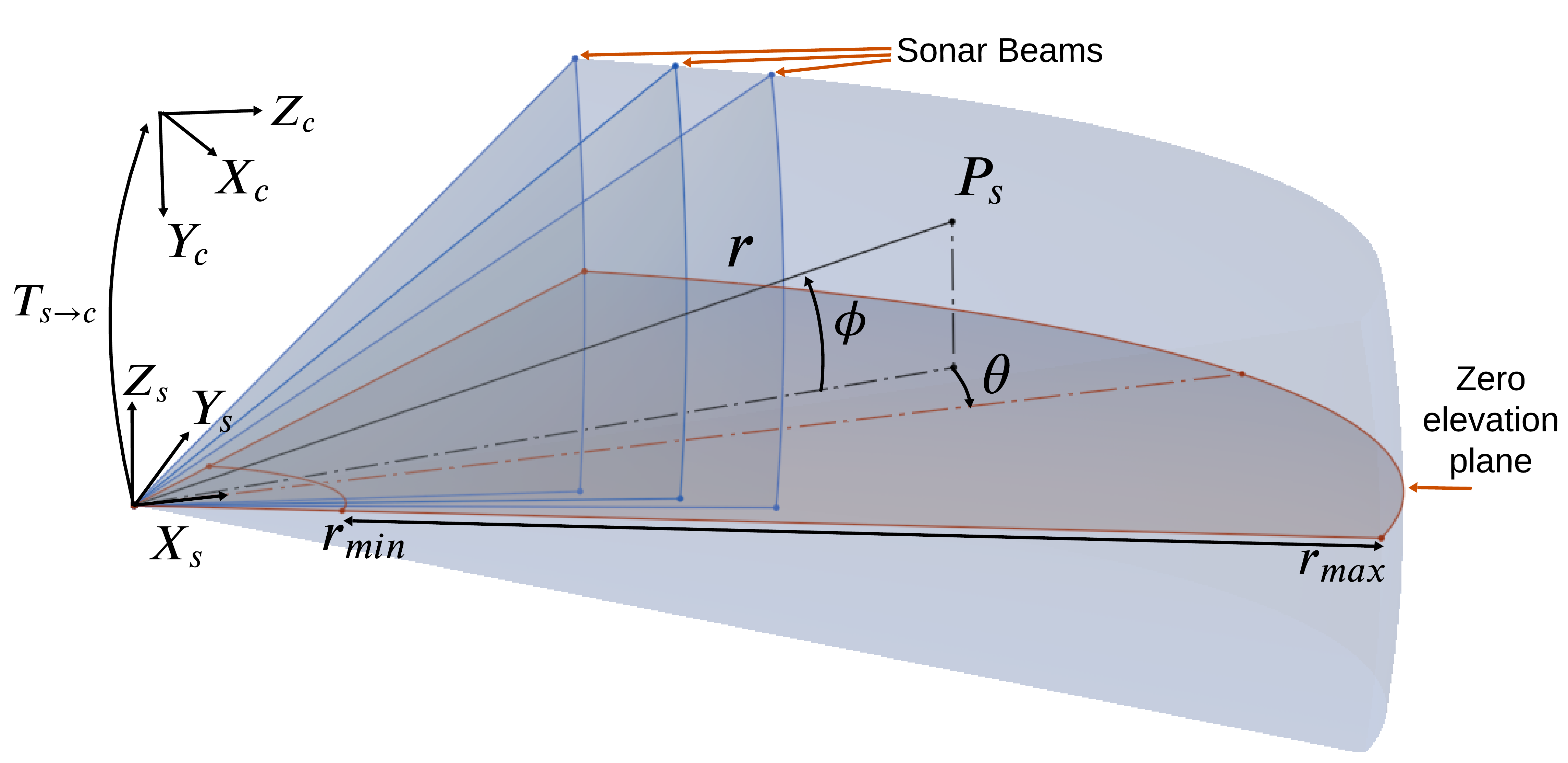}
    \caption{\textbf{Forward looking imaging sonar model.} The point $\mathbf{P}_s$ can be represented by $[r, \theta, \phi]^T$ in a spherical coordinate frame. The range $r$ and the bearing angle $\theta$ of $\mathbf{P}_s$ are measured, while the elevation angle $\phi$ is not captured in the resulting 2D sonar image. The 3D transformation $\mathbf{T}_{s \to c}$ from sonar to camera frame is also depicted.}
    \label{fig:FLS}
    \vspace{-4mm}
\end{figure}
 Imaging sonars sense a 3D volume, as illustrated in Fig. \ref{fig:FLS}. An imaging sonar measures points in spherical coordinates by emitting acoustic pulses and measuring the associated intensity $\beta \in \mathbb{R}_+$ from their returns. The transformation of a 3D point from spherical $[r, \theta, \phi]^T$ to Cartesian coordinates with respect to the sonar reference frame is given by
  \begin{equation}
    \mathbf{P}_s = 
    \begin{pmatrix}
        x_s\\
        y_s\\
        z_s
    \end{pmatrix}
    =r
    \begin{pmatrix}
        \cos\phi \cos\theta\\
        \cos\phi \sin\theta\\
        \sin\phi
    \end{pmatrix}.
    \label{eq:sonar3D}
\end{equation}
Here $r \in \mathbb{R}_+$ is the range, $\theta \in \Theta$ is the bearing, and $\phi \in \Phi$ is the elevation angle, with $\Theta, \Phi \subseteq [-\pi, \pi)$. Each acoustic ping return corresponds to a beam that spans elevation values $\phi \in [\phi_{\min}, \phi_{\max}]$, thus the sonar produces a 2D intensity image $I(r,\theta)$ that lacks explicit elevation information. 

\subsection{Camera Model}
Transforming a 3D point from the sonar reference frame $\mathbf{P}_s$ to the camera reference frame $\mathbf{P}_c$ is achieved using the extrinsic parameters $\mathbf{R}_{s \to c}$ the rotation matrix and $\mathbf{t}_{s \to c}$ the translation vector between the sonar and camera frames.
\begin{equation}
    \mathbf{P}_c = 
    \begin{pmatrix}
        x_c \\
        y_c \\
        z_c 
    \end{pmatrix}
    =\mathbf{R}_{s \to c}\mathbf{P}_s+\mathbf{t}_{s\to c}
    \label{eq:SonartoCamera}
\end{equation}
Using the pinhole camera model, a 3D point in the camera reference frame $\mathbf{P}_c$ is projected onto a 2D point $\mathbf{p}_c$ on the image plane using the intrinsic camera matrix $\mathbf{K}$.
\begin{equation}
    \mathbf{p}_c = \begin{pmatrix}
        u \\
        v \\
        1 
    \end{pmatrix} = \mathbf{K}\frac{\mathbf{P}_c}{z_c}, \qquad
      \mathbf{K} =
        \begin{pmatrix}
        f_x & 0 & c_u\\
        0 & f_y & c_v\\
        0 & 0 & 1
    \end{pmatrix}.
    \label{eq:3Dto2D}
\end{equation}
Here, $(f_x, f_y)$ are the focal lengths in pixels, and $(c_u, c_v)$ is the optical center of the camera.

\vspace{-1mm}
\subsection{Data Association Problem}
\vspace{-1mm}
We assume a robot equipped with two forward-looking imaging sonars and one forward-looking camera, mounted with overlapping FOVs (Fig. \ref{fig:FOVandROV}). This configuration yields regions observed by all three sensors as well as areas covered only by the camera and a single sonar. Proper calibration allows measurements in overlapping regions to be associated to the same physical location.

Recovery of 3D sonar points requires knowledge of the elevation angle $\phi$. Following \cite{McConnell2020StereoSonar}, features observed in orthogonal sonar images are fused in 3D as
\begin{equation}
\mathbf{\hat{P}}_s = \left( \frac{r^h + r^v}{2}, \theta^h, \theta^v \right),
\label{eq:StereoMeasurements}
\vspace{-1mm}
\end{equation}
where $r^h$, $r^v$ are range measurements from the horizontal and vertical sonars, and $\theta^h$, $\theta^v$ are the corresponding bearing angles. This association is only possible within the limited overlapping FOV, bounded by their vertical beam width. Outside this region, sonar measurements remain underconstrained and therefore undefined in 3D. Sonar provides range and bearing, while the camera contributes elevation. Fusing their complementary information enables full 3D reconstruction, but differing sensing mechanisms complicate integration and yield variable reliability. 

This paper explores how overlapping sensor information can be leveraged to maximize the number of potential 3D measurements. Moreover, since measurements differ in reliability, we also address how to map the environment while incorporating information with varying levels of certainty.

\vspace{-2mm}
\section{Proposed Approach} 
\label{sec:proposed_approach}
In this section, we present our framework for underwater sensing and mapping. We first exploit the overlapping sensor FOVs and apply matching procedures to maximize the number of fully defined 3D measurements, each with an associated reliability. These measurements are then fused in a volumetric mapping process that prioritizes the most reliable data. An overview of the system is shown in Fig. \ref{fig:pipeline}.

\vspace{-1mm}
\subsection{Sonar Feature Extraction}
\vspace{-1mm}
Not all pixels in an acoustic image \(I(r, \theta)\) contain meaningful information; noise and second returns are common. We therefore preprocess the data using the SOCA-CFAR filter \cite{El-Darymli2013SOCA}, a variant of the Constant False Alarm Rate (CFAR) technique \cite{Richards2005CFAR}, to detect salient features. Let  $F(r, \theta) = \text{SOCA-CFAR}(I(r, \theta))$ 
denote the filtered feature image, where only intensity values passing the CFAR threshold remain, \color{black}an example of this can be seen in Fig. \ref{fig:pipeline}\color{black}. This approach effectively mitigates second returns, which frequently appear in sonar imagery \cite{McConnell2020StereoSonar}. Unlike some prior works \cite{Collado2025}, all detected features in the sonar image are retained in $F$, even if there are multiple range returns with intensities exceeding the CFAR threshold for a given bearing angle.

\subsection{Optical Image Segmentation}
Instead of detecting individual point features in noisy, low-contrast underwater images, we propose detecting regions of interest (ROIs) corresponding to objects in the scene. Traditional segmentation methods often underperform underwater due to reliance on localized features and handcrafted extraction techniques, which are sensitive to poor contrast, color attenuation, and non-uniform lighting. Recent deep learning–based approaches have shown promising results in these conditions~\cite{Akram2025, Mohammadi2024}.

We adopt YOLO11n-seg~\cite{khanam2024yolov11} for its real-time performance, robustness, and multi-scale object detection. The model is initialized with COCO-pretrained weights and fine-tuned on a small hand-annotated dataset of representative \color{black}underwater \color{black} imagery (distinct from validation sequences). Data augmentation, including random flips, rotations, saturation, brightness, and exposure, expands the training set and improves robustness to blur, low light, and color distortion. An example segmentation is shown in Fig.~\ref{fig:pipeline}. The resulting segmented regions of interest are denoted as $R \subset \mathbb{R}^2$.

\subsection{Stereo Sonar Matching}
Given feature images $F_v$ and $F_h$ from the vertical and horizontal sonars, respectively, we need to associate features across the images to fully define them in 3D, as denoted in Eq. (\ref{eq:StereoMeasurements}). Similar to \cite{McConnell2020StereoSonar}, we perform range-based association, since range is the common denominator between the two sonars.

Unlike \cite{McConnell2020StereoSonar}, we do not consider intensity information when performing feature matching. Sonar return intensity depends on  several factors, including: reflectivity, incidence angle, and elevation ambiguity (i.e., the number of individual returns within the elevation angle of the beam.). According to the sonar image formation model formally presented in \cite{Qadri2023Neusis}, the intensity of each pixel in the sonar image is proportional to the cumulative acoustic energy reflected by all objects intersected by the acoustic arc at that range and bearing. Because our sonars are oriented ~90° apart, elevation effects vary significantly between them, making intensity unreliable for correspondence.

Consequently, matching is based solely on range and overlapping FOV. We only match features from the area inside the vertical \color{black}FOV $\Phi \in [\phi_{min}, \phi_{max}]$ \color{black}of each sonar’s orthogonal companion, shown in red and blue in Fig. \ref{fig:FOV}. Features at the same range are paired as:
\begin{equation}
    \mathcal{S}=\{(f_v,f_h)\mid r(f_v)=r(f_h), f_h,f_v \in \Phi\}.
\end{equation}
These matches are then transformed into a 3D point cloud:
\begin{equation}
    \mathcal{P}_{ss} = \{ \pi(f_v,f_h) \mid (f_v,f_h)\in\mathcal{S} \},
\end{equation}
where $\pi(\cdot)$ denotes the stereo \color{black}fusion \color{black} function defined in Eq. (\ref{eq:StereoMeasurements}). 

The overlap region shrinks with proximity, so nearby objects not directly in the robot’s 
line of sight
are likely missed. In addition, sonar measurements are inherently sparse and noisy, so small objects may not be reliably detected by both sonars simultaneously. Yet, objects close to the robot are often the most critical for navigation, as they represent imminent collision threats. 
To compensate, we store close-range features not matched stereoscopically (features closer than the minimum range among all valid stereo matches, $d_{\min}$):
\begin{equation}
\begin{aligned}
    \mathcal{C}_v = \{ f_v \in F_v \;\mid\; r(f_v) < d_{\min} \}, \\
    \mathcal{C}_h = \{ f_h \in F_h \;\mid\; r(f_h) < d_{\min} \}.
\end{aligned}
\end{equation} 
These sets of close-range, unmatched sonar features are later 
revisited, leveraging the higher-resolution information from the optical image (detailed in Secs.~\ref{sec:sonar_to_image}, \ref{sec:image_expansion}).

\subsection{Stereo Sonar to Image Projection}
$\mathcal{P}_{ss}$ is then projected onto the camera image using the intrinsic matrix $\mathbf{K}$ and the sonar-to-camera transform $\mathbf{T}_{s \to c}$: 
\begin{equation}
    \mathbf{p} =
    \begin{pmatrix}
        u \\
        v \\
        w 
    \end{pmatrix}/w
    = \mathbf{K} \mathbf{T}_{s \to c}
    \begin{pmatrix}
        \mathbf{P}_s \\
         1 \\ 
    \end{pmatrix} = \tau (\mathbf{P}_s),
\label{eq:Tau}
\end{equation}
where $\tau:\mathbb{R}^3 \to \mathbb{R}^2$ is the projection from 3D points in the sonar reference frame to 2D image pixels $\mathbf{p}\in \mathbb{R}^2$.

Next, the segmented regions of interest $R$ from the RGB image are applied as a mask, discarding all projected points outside $R$. This step helps to remove erroneous matches. The resulting point cloud is: $\mathcal{Q}_{ss}=\{ \mathbf{P}_s \in \mathcal{P}_{ss} \;\mid\; \tau(\mathbf{P}_s) \in R \}$.

If no region of interest is detected, we set $\mathcal{Q}_{ss} = \mathcal{P}_{ss}$ and skip 
the further inspection of close-range unmatched sonar features.
This ensures mapping continues with sonar-only data, which may be necessary in low-visibility conditions.

\subsection{Sonar to Image Projection}
\label{sec:sonar_to_image}
Due to elevation ambiguity in sonar data, each unmatched feature $f_h \in \mathcal{C}_h$ and $f_v \in \mathcal{C}_v$  potentially corresponds to multiple elevations. Inspired by \cite{Collado2025}, we project each feature to all possible 3D points:
\begin{equation}
    \mathcal{B}_j = \{\mathbf{P}_s\} = r
    \begin{pmatrix}
        \cos\phi_i \cos\theta\\
        \cos\phi_i \sin\theta\\
        \sin\phi_i
    \end{pmatrix},
    \quad \phi_i \in [\color{black}\phi_{min}, \phi_{max}\color{black}].
\end{equation}
Each 3D point set $\mathcal{B}_j$ is then projected onto the camera image using Eq. \ref{eq:Tau}.
Because of our sensor configuration, $\mathcal{B}_j$ projects to vertical lines for horizontal sonar features $f_h$, and to horizontal lines for vertical
features $f_v$. Let $\mathcal{O}$ represent the set of pixels $\mathbf{p}$ corresponding to the 3D point set $\mathcal{B}$ associated with the close-point feature sets $(\mathcal{C}_v \cup \mathcal{C}_h)$.

Finally, the segmented RGB regions $R$ are  applied as a mask, discarding all projections outside of $R$. The resulting point cloud is:
$\mathcal{Q}_s = \{ \mathbf{P}_s \in \mathcal{O} \;\mid\; \tau(\mathbf{P}_s) \in R \}$.

\subsection{Image Expansion}
\label{sec:image_expansion}
Imaging sonars have a limited and ambiguous vertical FOV. Similarly to \cite{Collado2025}, we expand this FOV using optical image information. Distance is derived from the original sonar measurement, while elevation is inferred from the optical image. 

For the horizontal sonar, let $\mathcal{O}_h$ be the set of optical image pixels corresponding to the 3D points $\mathcal{B}_h$ from the close feature set $\mathcal{C}_h$, restricted to the region of interest $R$. For each column of pixels in the optical image, we compute the mean depth $\bar{z}_c$ and assign it to all pixels in that column that lie within $R$ but outside the sonar’s vertical FOV (i.e., not in $\mathcal{O}_h$). \color{black}This expansion assumes the surface of the object in view has a fixed standoff distance from the sonar, which is common for man-made structures (i.e. piers), but may not apply to objects of arbitrary geometry\color{black}. These expanded 2D pixels $\mathbf{p}$ are then back-projected to 3D using:
\begin{equation}
    \mathbf{P}_c = \bar{z}_c \mathbf{K}^{-1} \mathbf{p}.
\end{equation}
The same process is applied to the vertical sonar, where $\mathcal{O}_v$ corresponds to $\mathcal{C}_v$ and expansion is performed along optical image rows instead of columns. The two resulting sets of 3D points are combined to form the expanded point cloud $Q_e$.

\subsection{Gaussian Process Occupancy Mapping}
We base our GP mapping solution on the fast and accurate 3D mapping framework presented in \cite{WangJinkun2016}, which has also demonstrated effectiveness in underwater sparse-data scenarios \cite{WangJinkun2019}. Standard GP mapping assumes a constant noise variance $\sigma_n^2$, treating all 
\color{black}observations \color{black}as equally reliable. Instead, we adopt a heteroscedastic GP, assigning each \color{black}observation \color{black}its own noise value $\epsilon$ that reflects the reliability of its source.
The GP predictive distribution at $\mathbf{x}_*$ remains Gaussian with
\begin{equation}
    \begin{aligned}
    \mu_* \;=\; k_*^\top\!\big(K + \Sigma\big)^{-1}\mathbf{y}, \\
    \sigma_*^2 \;=\; k(\mathbf{x}_*,\mathbf{x}_*) - k_*^\top\!\big(K + \Sigma\big)^{-1}k_*,
    \end{aligned}
\label{eq:pred_hetero}
\end{equation}
where $ \Sigma = \mathrm{diag}\!\big(\sigma_{n,1}^2,\dots,\sigma_{n,\color{black}N\color{black}}^2\big)$.
This formulation weights each 
\color{black}data point \color{black}by its own uncertainty, with larger $\sigma_{n,i}^2$ reducing that observation’s influence on the posterior.

In our framework, three point clouds are used, each with different reliability: 
$\mathcal{Q}_{ss}$ (stereo sonar) makes no geometric assumptions and is assigned high confidence $\alpha_{ss}$. $\mathcal{Q}_{s}$ (sonar-to-image projection) assumes sonar range applies within sonar vertical FOV, giving medium confidence $\alpha_s$. $\mathcal{Q}_{e}$ (image expansion) assumes sonar range applies across the optical elevation span. This makes the strongest assumption, receiving the lowest confidence $\alpha_e$. Finally, noise is set as $\sigma_{n,i}^2 = 1/\alpha_i$, linking GP uncertainty directly to measurement confidence.

\vspace{-1mm}
\section{Experiments and Results}
\label{sec:results}
To validate the proposed sensing and mapping framework, we evaluate its performance on diverse structures in both clear-water tank and turbid-water marina experiments.

\subsection{Hardware Overview}
The experimental platform is a heavy-configuration BlueROV2 (Fig.~\ref{fig:ROV}) equipped with a VectorNav VN-100 IMU (200 Hz), KVH DSP-1760 FOG (250 Hz), Bar30 pressure sensor (5 Hz), Rowe SeaPilot DVL (5 Hz), a low-light Sony Exmor IMX322/323 camera (5 Hz), and two forward-looking multibeam imaging sonars: Oculus M750d and M1200d. Both sonars were operated in their low-frequency wide-aperture modes (750 kHz and 1200 kHz), providing a $20^\circ$ vertical and $130^\circ$ horizontal FOV at a 3 m range.
\color{black}The sensor mounting configuration, shown in Fig.~\ref{fig:FOVandROV}, allows the use of measured sensor offsets as simple extrinsic calibration parameters, including a 10cm vertical offset between the sonar coordinate frames, as well as 15cm vertical and 15cm horizontal offsets between the sonar and camera frames.\color{black}

Onboard computation is provided by a Pixhawk, Raspberry Pi, and Jetson Nano running ROS and used for data logging to a topside computer. Algorithms are executed via real-time playback on a workstation (Titan RTX GPU, Intel i9 3.6 GHz CPU), representing hardware that could be embedded on an AUV. The vehicle trajectory is estimated by integrating DVL velocities, FOG and IMU angular rates, and pressure-based depth measurements. 

\vspace{-1mm}
\subsection{Benchmarks}
We compare our approach, Gaussian Process Confidence mapping with Stereo Sonar and RGB image (GPC SS RGB), against four baselines: (i) Standard GP occupancy mapping with Stereo Sonar and RGB (GP SS RGB), (ii) OctoMap with Stereo Sonar and RGB (Octo SS RGB), (iii) GP mapping with a single Sonar and RGB (GP S RGB) \cite{Collado2025}, and (iv) GP mapping with Stereo Sonar only (GP SS) \cite{McConnell2020StereoSonar}. The methods were chosen to showcase the impact of confidence values in the mapping process as well as the impact of using different sensor combinations. We exclude AONeuS \cite{Qadri2024} and Z-Splat \cite{Qu2024}, as they are not designed for real-time use and cannot generalize to scenes with unreliable visual information.

\vspace{-1mm}
\subsection{Indoor Tank Experiments} 
\begin{figure}[tb]
    \centering
    \includegraphics[width=1.0\linewidth]{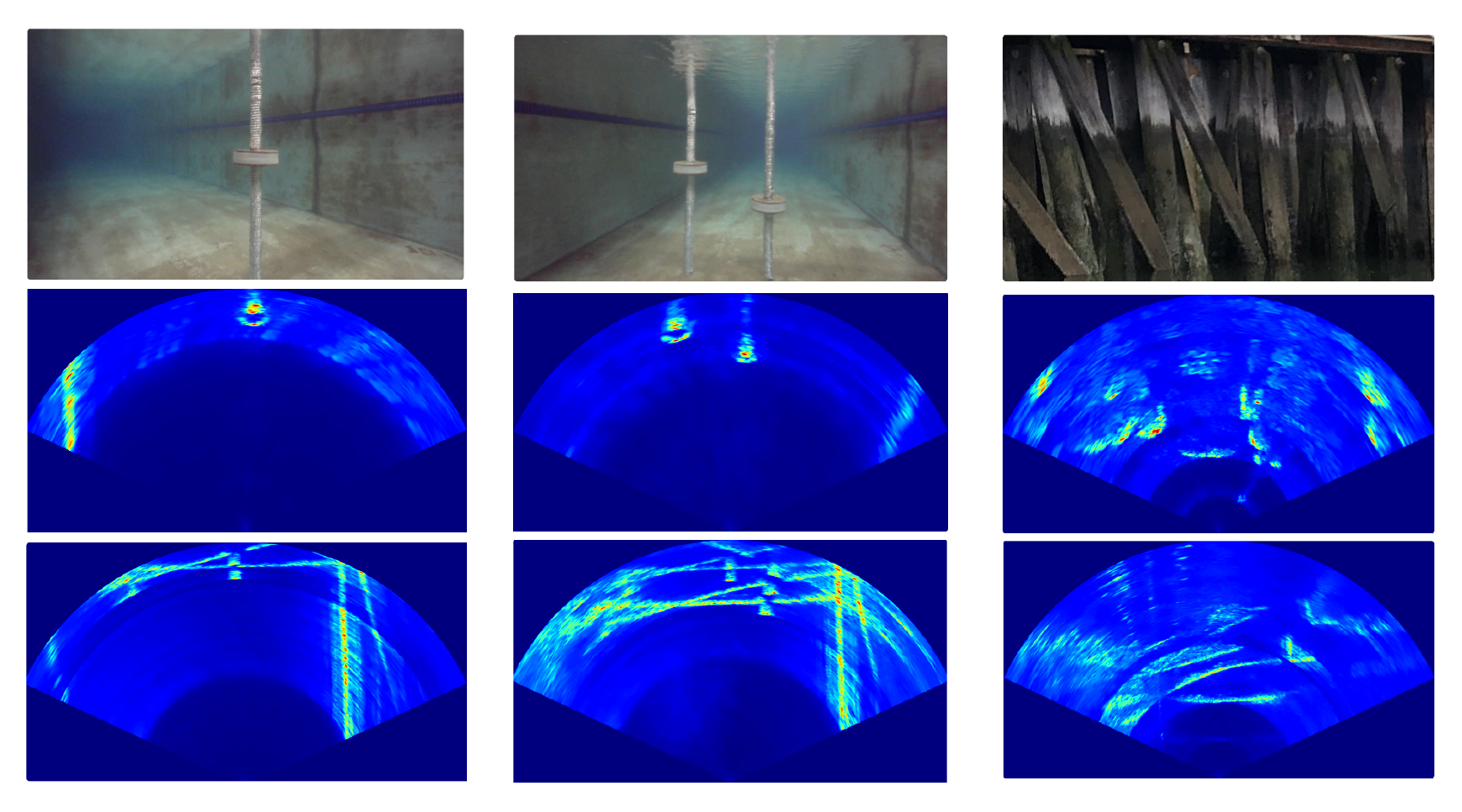}
    \caption{\textbf{Testing environments and representative sonar images.} Environments include: Tank Single Disk (left column), Tank Double Disk (middle column), and Marina Pier (right column). The top row shows an example image of each environment, the middle row shows an example horizontal sonar image, and the bottom row shows an example vertical sonar image.}
    \label{fig:FieldStructures}
    \vspace{-5mm}
\end{figure}
The testing structures were designed to be geometrically more complex than typical pier pilings (Fig. \ref{fig:FieldStructures}, left and middle columns), with surface discontinuities (sudden changes in dimensions) and few distinctive features, making feature matching and 3D reconstruction challenging. The first setup consists of a single pole-mounted disk, while the second includes two pole-mounted disks at different heights. A 200-image, hand-annotated dataset was used to train a pretrained YOLO11n-seg model for detecting pilings and disks.

Key frames were sampled every 5 cm or 10$^\circ$ for all mapping results. Voxel resolution was set to 2.5 cm. GP regression and classification hyperparameters were manually tuned: $\sigma_n^2 = 0.01$, \color{black}$l = 0.025$, $\sigma_f^2 = 0.1$, $\sigma_{min}^2 = 0.001$, $\sigma_t^2 = 50$\color{black}, and $\gamma = 100$. GPC confidence values for tank tests were set to $\alpha_{ss} = 20$, $\alpha_s = 1$, and $\alpha_e = 0.5$.

Error metrics were computed by comparing generated voxel maps against a ground-truth CAD model. Absolute error is defined as the shortest distance from each voxel center to the CAD model, with a bounding box used to exclude irrelevant objects \color{black} (e.g., tank walls, tank floor, etc.)\color{black}. Coverage was quantified via voxel count. Robot trajectories and full experiment playback are shown in the video attachment.

\subsubsection{\textbf{Tank Single Disk Results}}
\begin{table}[ht]
\centering
\begin{tabular}{|l|c|c|c|c|c|}
\hline
\textbf{Method} 
& \makecell{\textbf{MAE} \\ \textbf{(cm) ↓}} 
& \makecell{\textbf{RMSE} \\ \textbf{(cm) ↓}} 
& \makecell{\textbf{SD} \\ \textbf{(cm) ↓}} 
& \makecell{\textbf{Precision} \\ \textbf{(\%) ↑}} 
& \makecell{\textbf{Inlier} \\ \textbf{Voxels ↑}} \\
\hline
Octo SS RGB     & 5.9844           & 0.8106           & 0.5468     & 34.5989                & \textbf{1212}                 \\
GP SS RGB       & 6.6030           & 0.8816           & 0.5841     & 35.7711                & 450                  \\
\cellcolor{gray!30}GPC SS RGB      & \cellcolor{gray!30}\textbf{1.1712}          & \cellcolor{gray!30}\textbf{0.1579}         & \cellcolor{gray!30}\textbf{0.1059}   &\cellcolor{gray!30}\textbf{88.1844}               & \cellcolor{gray!30}306                  \\
GP S RGB        & 6.0007           & 0.8305           & 0.5741     & 38.8336                & 273                  \\
GP SS           & 1.3590           & 0.1819           & 0.1209     & 81.5789                & 31                   \\
\hline
\end{tabular}
\caption{Performance of volumetric mapping methods over the single disk structure in tank setting - proposed method is highlighted in gray. Best results per metric are shown in bold.}
\label{tab:single_disk_metrics}
\end{table}
\vspace{-4mm}
\begin{figure}[tbh]
    \centering
    \includegraphics[width=1\linewidth]{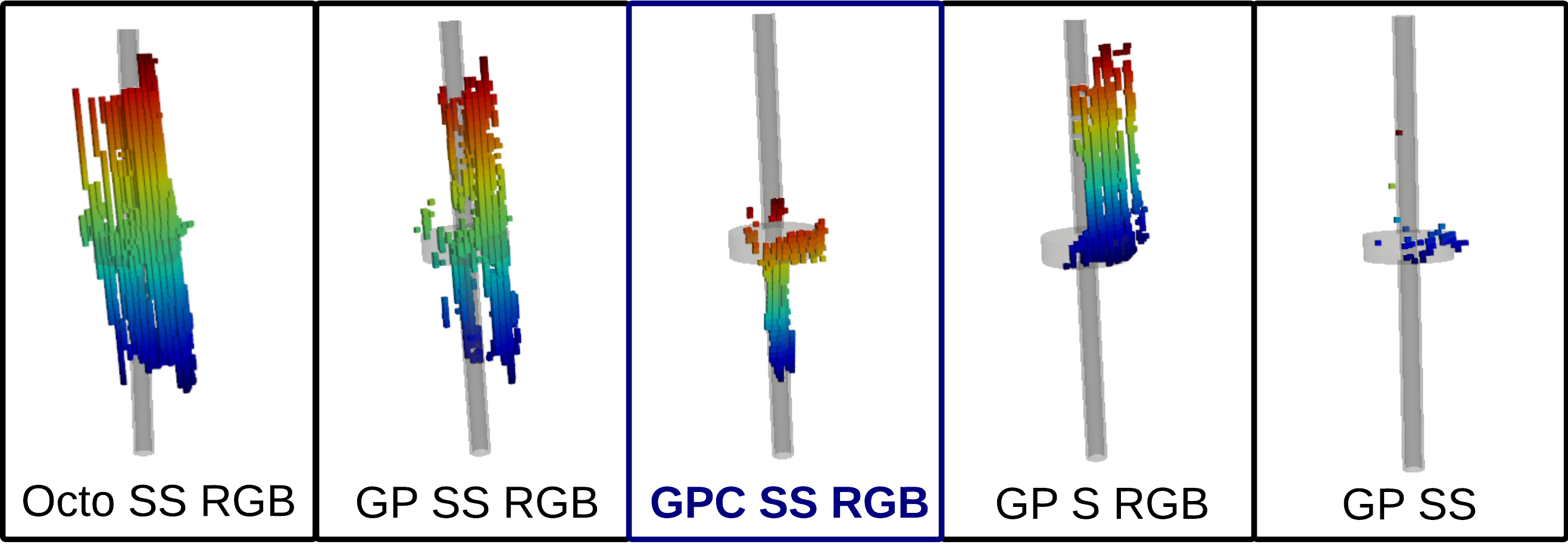}
    \caption{\textbf{Tank single disk mapping results.} Each column in the image shows the voxel map result for a different method. From left to right the order is: Octo SS RGB, GP SS RGB, \textbf{GPC SS RGB} (proposed approach highlighted in blue), GP S RGB, and GP SS. Voxel colors depict height.}
    \label{fig:single_disk_results}
    \vspace{-2mm}
\end{figure}
Numerical results are reported in Table \ref{tab:single_disk_metrics}, with visual results in Fig. \ref{fig:single_disk_results}. GP S RGB assumes a single distance from the robot applies to the entire structure. As a result, it reconstructs the scene as a cylinder with disk width, which is erroneous in this case where distances vary, leading to low accuracy. Octo SS RGB and GP SS RGB also show low accuracy, as they treat all data equally and cannot leverage confidence values to distinguish reliable measurements. GP SS achieves reasonable accuracy but has very limited coverage. In contrast, the proposed method, GPC SS RGB, attains the highest accuracy while maintaining substantial coverage.

\vspace{-1mm}
\subsubsection{\textbf{Tank Double Disk results}}
\begin{table}[ht]
\centering
\begin{tabular}{|l|c|c|c|c|c|}
\hline
\textbf{Method} & 
\makecell{\textbf{MAE} \\ \textbf{(cm) ↓}} & 
\makecell{\textbf{SD} \\ \textbf{(cm) ↓}} & 
\makecell{\textbf{RMSE} \\ \textbf{(cm) ↓}} & 
\makecell{\textbf{Precision} \\ \textbf{(\%) ↑}} & 
\makecell{\textbf{Inlier} \\ \textbf{Voxels ↑}} \\
\hline
Octo SS RGB & 2.6064 & 0.3141 & 0.4081 & 64.1379 & \textbf{1302} \\
GP SS RGB   & 2.5962 & 0.3338 & 0.4229 & 68.8243 & 521 \\
\cellcolor{gray!30}GPC SS RGB  & \cellcolor{gray!30}1.6219 & \cellcolor{gray!30}0.1651 & \cellcolor{gray!30}0.2315 & \cellcolor{gray!30}\textbf{81.5825} & \cellcolor{gray!30}598 \\
GP S RGB    & \textbf{1.3885} & \textbf{0.1306} & \textbf{0.1906} & 79.5385 & 517 \\
GP SS       & 2.3853 & 0.3511 & 0.4244 & 68.4685 & 76 \\
\hline
\end{tabular}
\caption{Performance of volumetric mapping methods over the double disk structures in tank setting - proposed method is highlighted in gray. Best results per metric are shown in bold.}
\label{tab:double_disk_metrics}
\end{table}
\begin{figure}[tbh]
    \centering    \includegraphics[width=1.0\linewidth]{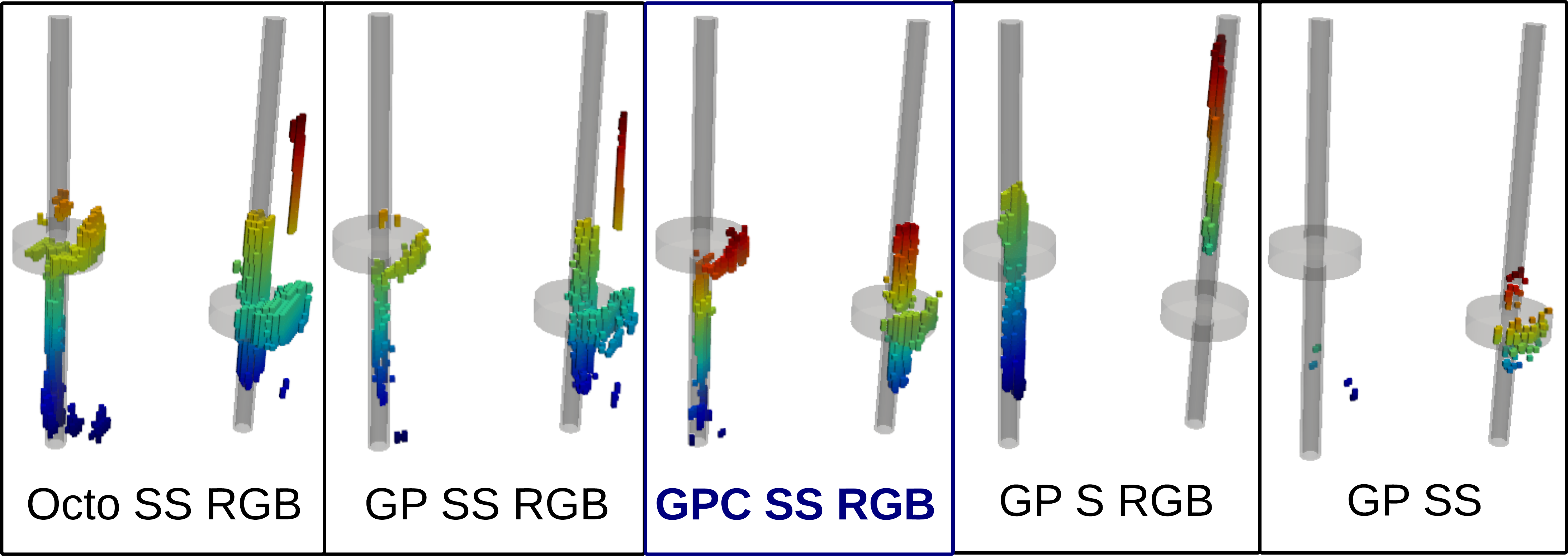}
    \caption{\textbf{Tank double disk mapping results.} Each column in the image shows the voxel map result for a different method. From left to right the order is: Octo SS RGB, GP SS RGB, \textbf{GPC SS RGB} (proposed approach highlighted in blue), GP S RGB, and GP SS. Voxel colors depict height.}
    \label{fig:double_disk_results}
    \vspace{-4mm}
\end{figure}
Numerical results are reported in Table \ref{tab:double_disk_metrics}, with visual results in Fig. \ref{fig:double_disk_results}. GP S RGB achieves the lowest errors, but its single-distance assumption causes it to capture the poles while completely missing the disks. GP SS still exhibits very limited coverage. Octo SS RGB and GP SS RGB capture some of the varying distance information but also introduce errors. Once again, our proposed method, GPC SS RGB, achieves the highest accuracy while maintaining substantial coverage. Overall, the proposed framework demonstrates superior capability to map complex geometries, prioritizing critical information for robotic decision-making and path planning.

\vspace{-1mm}
\subsection{Outdoor Field Experiments}
\begin{figure}[tbh]
    \centering
    \includegraphics[width=1\linewidth]{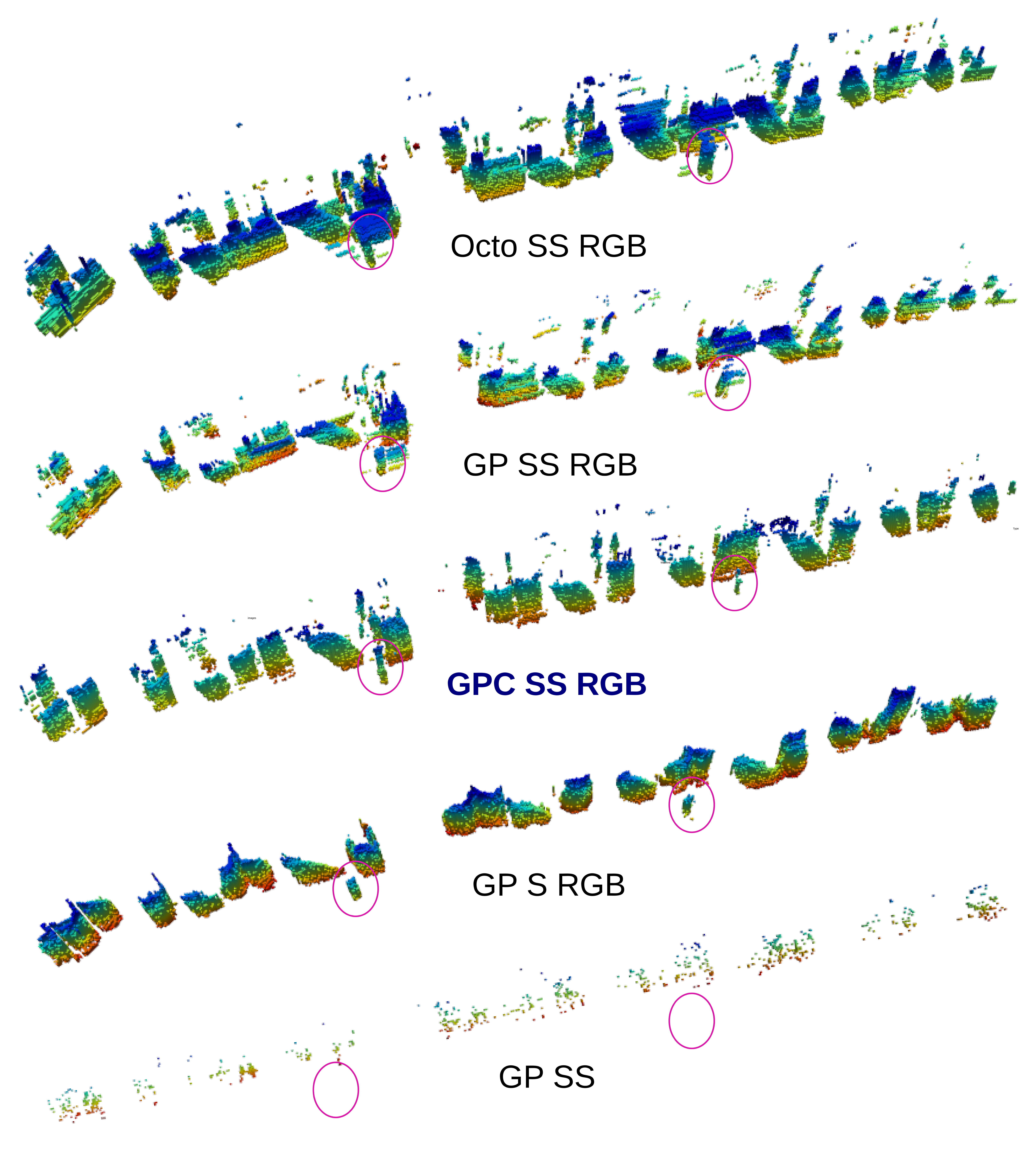}
    \caption{\textbf{Marina Pier mapping results.} Each row in the image shows voxel map results for a different method. From top to bottom the order is: Octo SS RGB, GP SS RGB, \textbf{GPC SS RGB} (proposed approach highlighted in blue), GP S RGB, and GP SS. Voxel colors depict height. Pink circles emphasize the presence of two small metal pipes in front of the wooden pilings.}
    \label{fig:field_results}
\vspace{-2mm}
\end{figure}

To evaluate our approach in the field, we deployed our robot 
\color{black}at the U.S. Merchant Marine Academy (USMMA) marina in King's Point, NY, a highly turbid environment, \color{black}
as can be seen in the example image in Fig. \ref{fig:pipeline}. 
The observed wood piling pier is shown in the right column of  Fig. \ref{fig:FieldStructures}. A 120-image hand-annotated dataset was used to train a pretrained YOLO11n-seg model to detect pilings. GPC confidence values were set to $\alpha_{ss} = 20$, $\alpha_s = 2$, and $\alpha_e = 1$. Due to the lack of ground-truth data, only qualitative results are presented in Fig. \ref{fig:field_results}.

The resulting pier voxel maps are shown in Fig. \ref{fig:field_results}. The pier includes two small pipes in front of the wooden pilings, circled in pink in the maps. GP SS is very sparse and fails to capture the pipes. GP S RGB produces a clean map with minimal outliers and captures the pipes perfectly, but it provides no information beyond the first sonar return at each range, and diagonal pier pilings appear triangular. GP SS RGB and Octo SS RGB clearly overestimate occupancy. Our method, GPC SS RGB, successfully captures the two small pipes, preserves the diagonally-oriented beams, and maps an inner row of pier pilings, even when partially occluded. As intended, the method prioritizes relevant information over raw coverage, capturing close objects essential for safety while maintaining background information important for path planning.

Mapping time results for GP, GPC, and Octomap across all experiments are reported in Table \ref{tab:time}. For fairness, only SS RGB methods are compared, since they map the same information. GPC and GP mapping show nearly identical computation times, indicating that adding confidence values has minimal impact. In contrast, Octomap has the highest computation times, in some cases more than double those of the GP-based approaches.
\vspace{-1mm}
\begin{table}[h!]
\centering
\setlength{\tabcolsep}{3pt} 
\centering
\begin{tabular}{|l|p{2.1cm}|p{2.2cm}|>{\centering\arraybackslash}p{2.0cm}|}
\hline
\textbf{Method} & \textbf{Tank Single Disk} & \textbf{Tank Double Disk} & \textbf{Marina Pier} \\
\hline
Octo SS RGB  & 0.1495 $\pm$ 0.0838 & 0.1168 $\pm$ 0.0414 & 0.2797 $\pm$ 0.3474 \\
GP SS RGB    & \textbf{0.0551} $\pm$ 0.0634 & \textbf{0.0699} $\pm$ 0.0363 & 0.1040 $\pm$ 0.0365 \\
\cellcolor{gray!30}GPC SS RGB   & \cellcolor{gray!30}0.0555 $\pm$ \textbf{0.0197} & \cellcolor{gray!30}0.0712 $\pm$ \textbf{0.0311} & \cellcolor{gray!30}\textbf{0.0970} $\pm$ \textbf{0.0324} \\
\hline
\end{tabular}
\caption{Mapping time (in seconds) with standard deviation. The proposed approach is highlighted in gray. Best-performing results are shown in
bold.}
\label{tab:time}
\vspace{-4mm}
\end{table}
\vspace{-2mm}
\section{Conclusion}
\vspace{-2mm}
\label{sec:conclusion}
This paper presents a sensor fusion and volumetric mapping framework for underwater robots designed to operate reliably under varying visibility to support safe navigation in cluttered environments. The approach leverages overlapping sonar and camera fields of view, assigns confidence values to each fully defined 3D measurement, and fuses them in a volumetric mapping framework that prioritizes the most reliable data. The proposed framework is optimized for challenging underwater environments by relying primarily on sonar, incorporating camera data only when available, while falling back to stereo sonar-only mode if visual information is unavailable. CNN-based segmentation enables efficient training via transfer learning and data augmentation on visual datasets, avoiding manual parameter tuning and dependence on sonar datasets. The sensor fusion prioritizes mapping close-range, critical features for safe navigation.

We evaluate the method against state-of-the-art techniques across scenes with varying geometric complexity. Our approach outperforms existing methods in accuracy while maintaining adequate scene coverage. Field experiments further demonstrate the system's ability to map both small and large structures in low-visibility conditions. While the framework demonstrates robustness and high accuracy, future work includes extending the method to 
\color{black}dynamic and large-scale environments, including improvements in sonar–camera calibration, and exploring adaptive sensing under different visibility conditions\color{black}. 
\vspace{-4mm}

\bibliographystyle{IEEEtran}
\bibliography{bib.bib}
\end{document}